# Classification and Recognition of Encrypted EEG Data Based on Neural Network

Yongshuang Liu[1,2], Haiping Huang[1,2], Fu Xiao[1,2], Reza Malekian[3], Wenming Wang[1,2,4]

1. College of Computer, Nanjing University of Posts and Telecommunications, Nanjing 210023, China

2. High Technology Research Key Laboratory of Wireless Sensor Network of Jiangsu Province, Nanjing 210023, China

3. Department of Computer Science and Media Technology, Malmö University, Malmö, 20506, Sweden

4. School of Computer and Information, Anqing Normal University, Anqing 246011, Anhui, China

hhp@njupt.edu.cn (H. Huang), reza.malekian@ieee.org (R.Malekian)

**Abstract:** With the rapid development of Machine Learning technology applied in electroencephalography (EEG) signals, Brain-Computer Interface (BCI) has emerged as a novel and convenient human-computer interaction for smart home, intelligent medical and other Internet of Things (IoT) scenarios. However, security issues such as sensitive information disclosure and unauthorized operations have not received sufficient concerns. There are still some defects with the existing solutions to encrypted EEG data such as low accuracy, high time complexity or slow processing speed. For this reason, a classification and recognition method of encrypted EEG data based on neural network is proposed, which adopts Paillier encryption algorithm to encrypt EEG data and meanwhile resolves the problem of floating point operations. In addition, it improves traditional feed-forward neural network (FNN) by using the approximate function instead of activation function and realizes multi-classification of encrypted EEG data. Extensive experiments are conducted to explore the effect of several metrics (such as the hidden neuron size and the learning rate updated by improved simulated annealing algorithm) on the recognition results. Followed by security and time cost analysis, the proposed model and approach are validated and evaluated on public EEG datasets provided by PhysioNet, BCI Competition IV and EPILEPSIAE. The experimental results show that our proposal has the satisfactory accuracy, efficiency and feasibility compared with other solutions.

## 1 Introduction

The rapid technological convergence of brain science, bionics and artificial intelligence, has enabled EEG-based information processing and control to emerge as a promising industrial application domain that has significant potential to improve the life quality of the elderly and the disabled. EEG signals are electrical signals generated by biological brain activity of human. Elderly and disabled people can wear specific devices that collects EEG data to generate control signals through motor imagination, so that they can remotely operate household appliances and other facilities with ideas. A large number of studies have confirmed that EEG signals have different characteristics in different states, which can be identified and classified by machine learning methods. As a channel established between the brain and external devices to transmit information, BCI collects and extracts EEG signals generated by the brain to recognize the human mind, so as to achieve message transmission and instruction control [1-3]. Therefore, most of researchers are more concerned with how to use machine learning methods to realize the applications of EEG signals and brain-computer interfaces, such as smart home [4], intelligent medical [5], etc., but few people pay attention to the privacy and security of EEG data itself.

It is assumed that a specific wearable EEG device, for example, Emotiv Insight, is used to collect the EEG data of one user through motor imagination in an IoT application scenario. The collected data is sent to the corresponding terminal for classification and recognition, and then subsequent operations are performed according to the recognition result, such as controlling TV program switching and adjusting the air conditioner temperature and so on. However, the scenario mentioned above is built on the premise that there are no malicious attackers. Actually, such an EEG-based IoT application system is vulnerable, and the attacker can easily achieved by launching passive and active attacks: 1) Passive attack means that the attacker can listen to the channel and obtain the data transmitted in the channel; 2) Active attack means that the attacker can forge the EEG data or tamper with the EEG data sent by the user, so as to release the command to the device and complete the operation required by the attacker. Furthermore, we introduce another application scenario that if a hospital wants to use the patient's EEG data to predict the possibility of epilepsy, but this work may be left to the prediction service provider (such as cloud service). Therefore, this process may reveal patient-related information. In such a situation, we not only require accurate prediction capabilities, but also pay attention to ensuring the privacy and security of the data. Both of the above cases have exposed some serious problems, that is, the security and privacy of user data cannot be guaranteed. Under general circumstances, when the training of the neural network model is completed, the user can adopt the model to make prediction tasks, however the sample imported by the user is the original data, which will generate potential security and privacy risks. Therefore, how to effectively handle encrypted EEG data has become an urgent problem to be solved.

Aiming at the security problem mentioned above, some researchers have started to focus on the application of homomorphic encryption algorithms to the field of machine learning. Although using homomorphic encryption to process EEG data can effectively curb various attacks, there are still exist some challenging problems. Firstly, the full homomorphic encryption algorithm is the most ideal choice because it supports both multiplication and addition operations, but its high time overhead causes there is a large gap from the practical application. Secondly, although the semi-homomorphic encryption algorithm is much lower in time complexity and implementation difficulty than the full homomorphic one, most of them cannot directly deal with the decimal problem. Thirdly, the classifiers used in the existing solutions may not be applicable because most researchers use complex artificial feature extraction methods, and the instability of these methods would probably further affect the accuracy of encrypted EEG signals classification.

The above challenging problems motivate us to design an effective scheme distinguished from current solutions, which takes classification accuracy, implementation efficiency and security & privacy into account. Therefore, our contributions in this paper can be summarized as follows:

(1) To ensure the privacy and security of EEG data in practical applications, the semi-homomorphic encryption algorithm is adopted to encrypt and decrypt EEG data for the purpose of low time cost, and meanwhile the data is converted to integers through appropriate scaling, which effectively solves the decimal problem in the encryption process.

(2) To improve the accuracy and stability of encrypted EEG data classification, the feedforward neural network is used as the classifier, which does not require artificial feature extraction. Furthermore, in order to enable the feedforward neural network to better handle encrypted data, the approximate function is employed instead of the activation function, so that it can still maintain a certain homomorphism after processing the encrypted data.

(3) Through hyperparameters adjustment such as the size of hidden layer neurons and the learning rate in the simulated annealing algorithm, the best parameters are achieved for the purpose of keeping a balance between the precision and time consumption. In addition, the proposed solutions are analyzed and verified from the perspective of time complexity and security, and the performance is evaluated by the confusion matrix and F1-score. Compared with other solutions, the feasibility and effectiveness of our proposal is demonstrated.

The rest of this paper is organized as follows: section 2 reviews the relevant research work; and section 3 describes the classification method of encrypted EEG data based on neural network, which includes the improvement of homomorphic encryption algorithm and neural network model; the experimental analysis and discussion are illustrated in section 4; finally, section 5 concludes the whole paper.

## 2 Related work

Over the last decade, much attention has been drawn to EEG data modeling and classification. Yong Jiao et al. [6] introduced a novel sparse group representation model (SGRM) and the corresponding algorithm to exploit inter-subject information for constructing an efficient classification model in motor imagery-based BCI applications. By using two public motor imagery-related EEG datasets, the extensive experimental comparisons were carried out between the proposed method and two other state-of-the-art methods. Their approach obtained the accuracy of 78.2%. Chatterjee R et al. [7] put forward fuzzy discernibility matrix (FDM) to find the best feature subset of EEG signals and use the Support Vector Machine (SVM) and Ensemble variants of classifiers to evaluate the performances of the selected feature subsets. But the accuracy of EEG signals recognition based on SVM is not desirable especially when dealing with large volume EEG data and multi-classification problem. In [8-9], the method based on deep convolutional neural network was proposed to classify EEG data. The accuracy of the results was 86.41% in [8] and 85.62% in [9], respectively. Nienhold D et al. [10] used genetic algorithm and feedforward neural network to classify and identify EEG data, and the error rate was reduced by about 10%. Alomari et al. [11] used Coiflets wavelet to analyze and process the EEG signals, and used amplitude estimation to extract the features, and finally employed machine learning algorithms for classification and recognition. Shenoy et al. [12] presented a shrinkage parameter based regularized approach for Filter Bank Common Spatial Pattern. This method was not only computationally tractable, but also overcome the defects of heuristic approach of traditional cross validation based parameter tuning. Ward et al. [13] used I-vectors and Joint Factor Analysis to extract the features, and meanwhile employed universal background models (UBMs) for classification. However, the approaches in [11-13] have not achieved the satisfactory accuracy. From the above, it can be seen that neural network algorithms have higher accuracy in the classification and recognition of EEG data than other machine learning related algorithms, and they do not require the feature extraction and are more suitable for processing huge data sets.

Recently, some scholars have completed a small amount of research on how to classify and identify EEG data after encryption. Kumar P et al. [14] proposed an approach which includes both addition and multiplicative homomorphic encryption algorithms. First, they encrypted the EEG data with the additive homomorphic encryption algorithm; second, when the multiplication operation is called, the corresponding ciphertext would be decrypted and encrypted again using a multiplicative homomorphic algorithm; finally the result would be decrypted and again encrypted using the additive homomorphic algorithm. However, the complexity of this approach is much larger than a single homomorphic encryption scheme. Clifton C et al. [15] employed a secure multi-party computing model to achieve data privacy protection. Chabanne H et al. [16] used a combination of neural network and full homomorphic encryption to establish a 6-layer model for the MNIST database to verify the reliability of the method. Gilad et al. [17] presented a method to convert learned neural networks to CryptoNets, which can be applied to encrypted data. Wu et al. [18] proposed a scheme that achieved database security, DO's key confidentiality, query privacy and the hiding of data access patterns. Their scheme efficiently solved the problem of privacy preserving KNN classification over the encrypted cloud database. Marcano et al. [19] completed an implementation of Fully Homomorphic Encryption (FHE) with Convolutional Neural Networks (CNN) for privacy-preserving. The combination of FHE and CNN was achieved by using a logic circuit that could perform deep learning algorithms on ciphertext rather than plaintext. Wang et al. [20] proposed a novel fully homomorphic Extreme Learning Machine (ELM), which enabled cloud searching tasks under a fully protected environment without compromising the privacy of users. However, the full homomorphic encryption requires relatively high time complexity. Therefore, finding low-overhead and high-precision methods for encrypting EEG data remains a challenging problem.

## 3 The Proposed Approach

The symbols in this paper are shown in table 1, which will be used in the following sections.

**Table1.** Symbol and Description

| Symbol | Description |
|---|---|
| $D$ | decryption operation |
| $E$ | encryption operation |
| $a,b$ | plaintext |
| $gcd$ | greatest common divisor |
| $n, g$ | public key |
| $\lambda, \mu$ | private key |
| $c$ | ciphertext |
| $z$ | the value of the node |
| $E$ | the value of loss function |
| $y_k$ | $k$th predicted value |
| $t_k$ | $k$th actual value |
| $r_{xy}$ | correlation coefficient |
| $m$ | the number of hidden layer neurons |
| $n_i$ | the number of nodes in the input layer |
| $n_o$ | the number of nodes in the output layer |
| $a$ | constant |
| $precision$ | the proportion of all positive predictions that are correctly predicted |
| $recall$ | the proportion of all real positive observations that are correctly predicted |
| $TP$ | true positive |
| $FP$ | false positive |
| $FN$ | false negative |
| $F_1$-Score | the weighted average of the precision and recall |

In this section, we first show the flow chart of the entire scheme. The system structure is shown in Figure 1, which consists of two parts: offline experiment and online experiment.

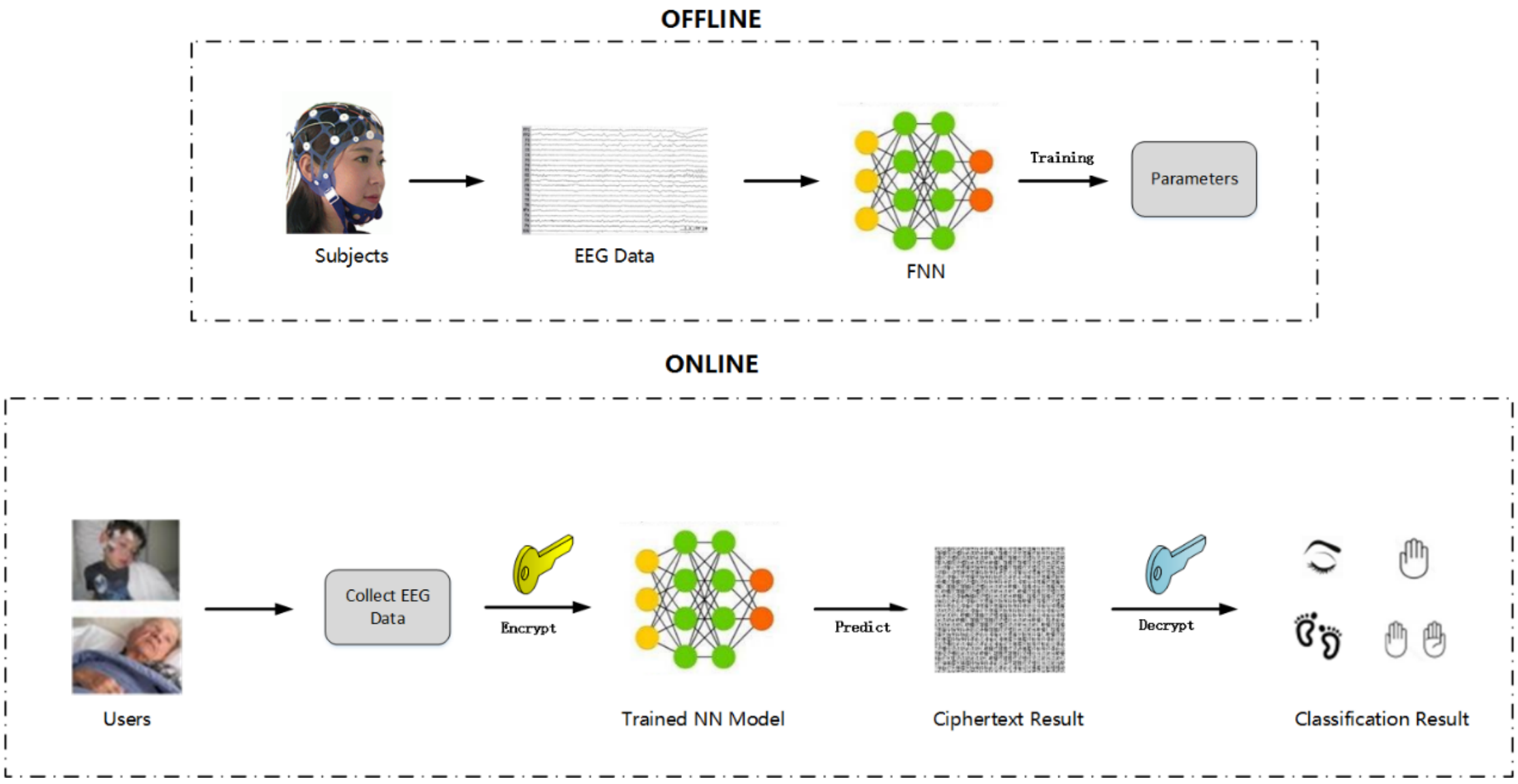

**Fig. 1.** Flow Chart of the Proposed Approach

In the offline experiments, we use unencrypted EEG data to train neural networks models (described in Section 3.2) and continually adjust hyperparameters (described in Section 4.3) to seek the highest accuracy, and then we determine the hyperparameters of the best model. In the online experiments, we use the public key to encrypt the untrained data in the EEG data set, and then put them into the trained neural network by the offline experiments for classification. After the classification is achieved, since the output result is also ciphertext format, the key owner can decrypt the ciphertext to obtain the corresponding classification result.

### 3.1 Homomorphic Encryption and the Solution to Decimals Problem

Homomorphic encryption is a key technology of our proposal. The so-called homomorphic encryption algorithm refers to an encryption algorithm with the following properties: for different data encrypted with the same key, an operation mode can be constructed so that this operation performed on the ciphertext can be mapped to a linear operation on the plaintext. We assume that the encryption operation of the algorithm is *E*, the decryption operation is *D*, and there are two plaintexts *a* and *b*. If an operation ($\oplus$) on a ciphertext domain can be constructed as

$$D(E(a \oplus b)) = a + b \tag{1}$$

then we call the encryption algorithm as additive homomorphic encryption algorithm. Similarly, if an operation ($\otimes$) can be constructed on a ciphertext domain as

$$D(E(a \otimes b)) = a \times b \tag{2}$$

then the encryption algorithm is called as multiplicative homomorphism encryption algorithm.

An encryption algorithm that satisfies the homomorphism of partial linear operations is called a semi-homomorphic encryption algorithm. The Paillier algorithm [21] is a semi-homomorphic encryption algorithm that satisfies the characteristics of the addition homomorphism. The encryption scheme randomly selects two large primes $p$ and $q$, and the product of them is $n$. $\lambda$ is the least common multiple of $p$-1 and $q$-1, and $g$ is an integer randomly selected from $Z_{n^2}^*$ that satisfies

$$gcd\left(L\left(g^{\lambda} \bmod n^2\right), n\right) = 1 \tag{3}$$

where $L(x) = (x-1)/n$, $\mu = L\left(g^{\lambda} \bmod n^2\right)^{-1} \bmod n$, *gcd* is the abbreviation of greatest common divisor.

After the above operation, the public key is ($n$, $g$) and the private key is ($\lambda$, $\mu$).

A message $m$ is encrypted by computing

$$c = g^m \cdot r^n \mathrm{mod} n^2 \tag{4}$$

where $r \in Z_{n^2}^*$, $0 \le m < n$ and $0 < r < n$. Decrypting is done by computing the following equation

$$m = L\left(c^\lambda \mathrm{mod} n^2\right) \cdot \mu \mathrm{mod} n \tag{5}$$

where $c$ is the ciphertext to be decrypted, $\lambda$ and $\mu$ are obtained by the private key.

**The solution to decimals problem:** The Paillier algorithm is adopted which is based on the difficult problem of compound residual classes. However, the numbers involved in number theory are usually integers, and how to process decimals from EEG data is a challenging issue, the Paillier algorithm cannot directly encrypt decimals. Actually, most encryption algorithms cannot handle decimals directly. For the problem of dealing with decimals, the encoding method is employed. This method can convert real numbers to fixed-precision numbers and then use their binary representation to convert them to a polynomial with coefficients given by binary expansion. When this polynomial takes a value at 2, it will return the encoded value. In fact, it is a high-precision process of converting decimals to corresponding integers. After the encoding is finished, the corresponding integers will be encrypted. Similarly, after the ciphertext is decrypted, the corresponding plaintext needs to be decoded to obtain the final result. In our experiments, keeping 5-bit to 10-bit accuracy of the input and the network weight is sufficient to ensure the accuracy of the neural network.

### 3.2 Neural Networks and the Use of Approximation Function

The purpose of this section is to demonstrate the application of neural networks in encrypted EEG data. In our work, we focus on fully connected feed-forward neural networks. The feed-forward neural networks includes an input layer, a hidden layer and an output layer. The data sequentially passes through the neurons of each layer in the above order, and each neuron receives only the output of the neurons in the upper layer, and there is no feedback between the networks.

Several common functions will be used in the neural networks. We have listed some of them here:

- *Weighted-Sum*: Multiply the vector of values of the next layer by the vector of weights and then plus the offset of the corresponding layer.
- *Activation Function*: Use sigmoid function and take the value of one of the nodes in the feeding layer to evaluate the function $z \leftarrow 1/(1 + exp(-z))$.
- *Loss function*: Use the cross entropy function to quantify the gap between the results of the neural networks and the correct results $E \leftarrow -\sum_k t_k log y_k$.

Since the Paillier algorithm is an addition homomorphism, it only supports addition operations. Among them, the weighted-sum function can be directly implemented since it uses only additions and multiplications, and the multiplications here are performed between the pre-calculated weights and the values of the input layer. Since the weights are not encrypted, they are essentially additions operations. Some networks will add a bias term after the weighted-sum, which is also a common addition operations. Moreover, the loss function is only used in the training phase and does not participate in the prediction process, so the loss function does not need to consider whether it is suitable for homomorphic encryption operations.

The sigmoid function is a nonlinear function, Xie et al. [22] replaced the activation function in neural networks with approximation polynomials and analyzed the performance of neural networks. The polynomial function can be seen as a multiplication operation, while the Paillier algorithm does not support multiplication homomorphism. Therefore, we use SPSS data analysis software to perform linear regression on the value of sigmoid function in the interval [-1, 1] with little precision loss. Table 2 shows the SPSS regression results report.

**Table 2.** Regression results report.

| Model | Non-standardized coefficient | | Standardized coefficient | t | Significant |
|---|---|---|---|---|---|
| | B | Standard error | Beta | | |
| 1(constant) | 0.500 | 0.000 | | 2456.965 | 0.000 |
| $x$ | 0.238 | 0.000 | 1.000 | 679.213 | 0.000 |

We can get the approximate formula of the activation function from the report.

$$f(x) = 0.238x + 0.5 \tag{6}$$

It can be seen that the significant of coefficients and constants given by SPSS are less than 0.001, that is, from a probabilistic point of view, 99.9% of the judgments are made to be convinced that the data satisfies the above approximation function. To further verify the accuracy of the approximation function obtained by regression, Figure 2 shows the error curve between the approximation function and the original function.

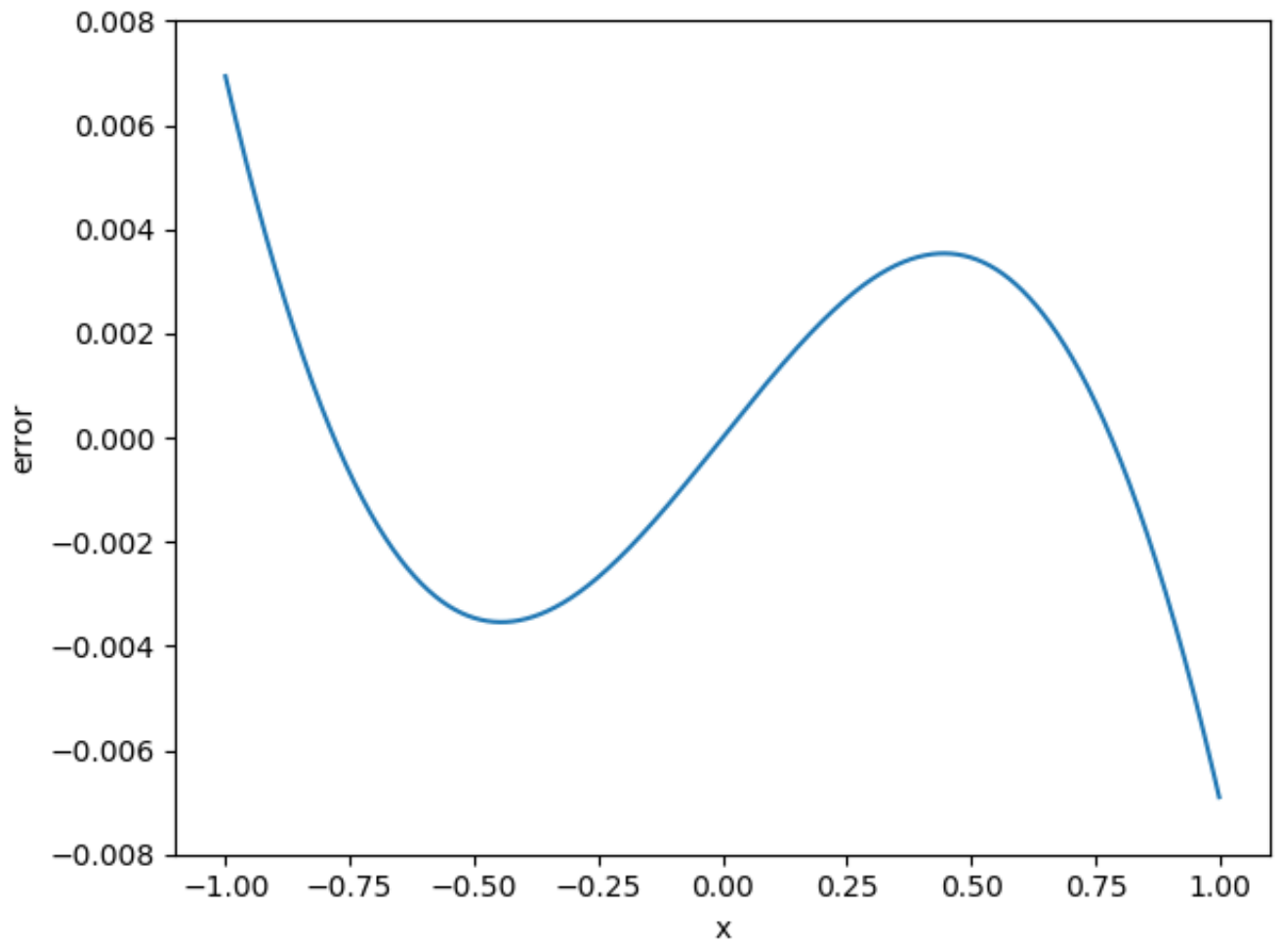

**Fig. 2.** The error curve between the approximation function and the original function.

It can be seen from figure 2 that the maximum absolute error of the two functions is around 0.006 in this interval, and when the absolute value of the data is around 0.5 (in fact, most of the data after normalization is near this value), the absolute error decreases within 0.004. We believe that this error can be tolerated when training neural networks.

# 4 Experimental Analysis and Evaluation

## 4.1 Data Set

In order to verify the validity of the proposal in this paper, the EEG testing data from PhysioNet eegmmidb database (EEG motor movement/imagery database) is employed. This data set was collected by the BCI2000 instrumentation system [23], which owns 64 channels with the EEG data sampling rate of 160 Hz. During the data collection of this database, the subject sits in front of one screen and performs the corresponding action as one target appears in different edges of the screen. Each subject performed 14 experimental running: the first two experiments of eyes opened and eyes closed for one minute each, and then four different tasks repeated three times (total 12 tests) and totally two minutes for one time.

Figure 3 is an example of visualization of EEG motion imaging signals.

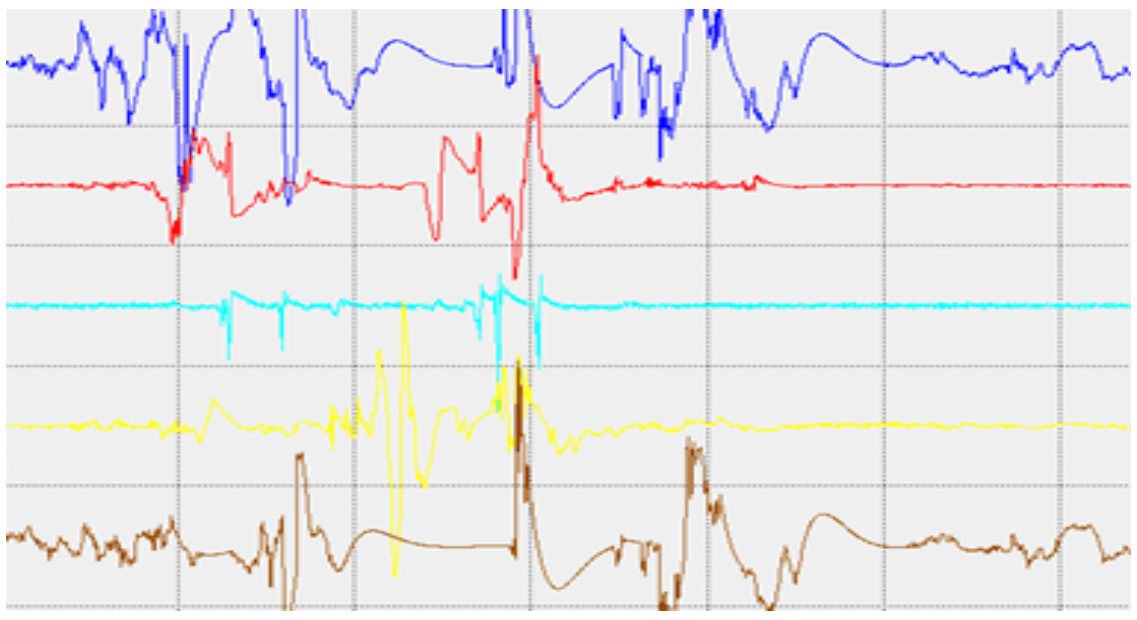

**Fig. 3.** EEG motion imaging signals

In this experiment, we selected four tasks as follows:

*Task* 1: The subject *opens his or her eyes* and keeps relaxed.

*Task* 2: The subject *closes his or her eyes* and keeps relaxed.

*Task* 3: A target appears on the top of the screen, and the subject focuses on *both hands* until the target disappears.

*Task* 4: A target appears on the bottom of the screen, and the subject focuses on *both feet* until the target disappears.

Specifically, we select 12800 EEG samples from 4 subjects (3200 samples for each subject) for our experiment. Every sample is one vector which includes 64 elements corresponding to 64 channels. Each sample corresponds to one task (from task 1 to task 4 separately is eye opened, eye closed, focus on both hands and focus on both feet). Every task is labeled as one class, and there are totally 4 class labels (from 1 to 4). The corresponding relationship is shown in table 3.

**Table 3.** Subject's movements and corresponding training labels in this experiment

| **Subject action** | **eye opened** | **eye closed** | **both hands** | **both feet** |
|---|---|---|---|---|
| Training label | 1 | 2 | 3 | 4 |

## 4.2 Data Preprocessing

The Pearson correlation coefficient, also known as the Pearson product-moment correlation coefficient, is a linear correlation coefficient. The Pearson correlation coefficient is used to reflect the statistics of the linear correlation between the two variables.

$$r_{xy} = \frac{\sum(X-\bar{X})(Y-\bar{Y})}{\left(\sqrt{\sum_{i=1}^{n}(X_i-\bar{X})^2}\right)\left(\sqrt{\sum_{i=1}^{n}(Y_i-\bar{Y})^2}\right)} \tag{7}$$

where the correlation coefficient is represented by $r$, $n$ is the sample size, $X$, $\bar{X}$ and $Y$, $\bar{Y}$ are the observations and mean values of the two variables respectively; $r$ describes the degree of linear correlation between two variables. The larger the absolute value of $r$ is, the higher the correlation degree is. The correlation coefficient ranges from -1 to +1. If $r>0$, it indicates that the two variables are positively correlated. If $r<0$, it indicates the two variables are negatively correlated.

By using the corr() method in scikit-learn, we can easily calculate the standard correlation coefficient between each pair of attributes. We use the sort_values() method to calculate the correlation degree between each attribute and the classification label. We select the data of the top 44 electrode channels with the highest correlation degree, such as Cpz, Oz, C4, Po4, F1, etc. Compared to 64 channels, 44 channels have less runtime and the accuracy is just a little loss. Therefore, the use of 44 channels of EEG data is feasible. As shown in Figure 4 (black electrode channels are deprecated, and the remaining electrode channels are used in the experiment).

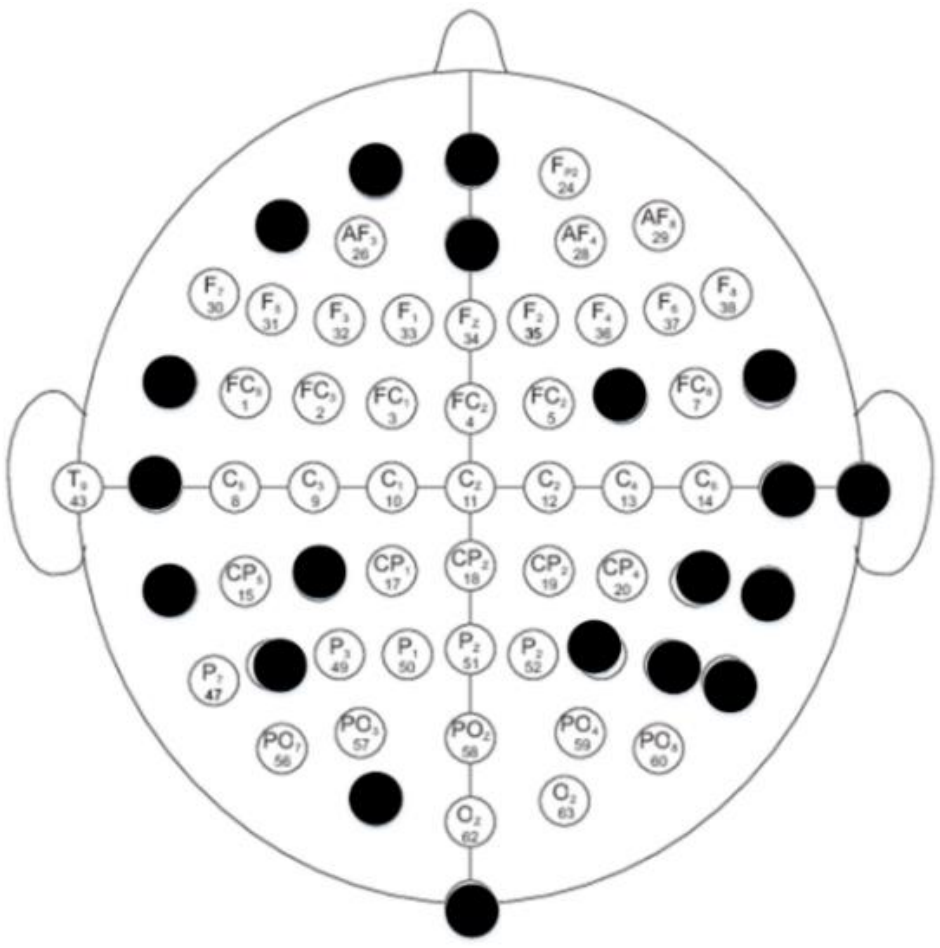

**Fig. 4.** Selection of electrode channels for EEG

### 4.3 Parameter Adjustment

In the experiment, the feed-forward neural network has three layers: the size of the input layer is 44, and 44 electrode values of the EEG data are successively input after data preprocessing; the size of the hidden layer is determined by the hyperparameter adjustment; the size of the output layer is 4, representing four classification results.

The following describes the setting of two hyperparameters that have a great influence on the neural network: one is the number of hidden layer neurons $m$ and the other is the learning rate $\eta$. And other hyperparameters in this experiment, such as the size of mini-batch, are selected according to our experience.

Determining the number of hidden layer neurons is an important task of neural network parameter adjustment. Too few neurons in the hidden layer will reduce the accuracy of the whole neural network, and too many will easily lead to over-fitting. Since the homomorphic encryption used in this paper is a relatively time-consuming task, each additional hidden layer of neurons will increase the encryption work time. Therefore, we need to consider the trade-off between the accuracy and the time consumption so that when determining the number of hidden layer neurons, we can reduce the time consumption within a certain accuracy tolerance. For this reason, it conducts several experiments separately by varying the number of neurons in the hidden layer when the other parameters are optimal, and meanwhile observes the accuracy of the best training set and the time required for the encryption neural network classification process. Each experiment is performed three times independently, and the experimental results are averaged and recorded in table 4.

**Table 4.** The relationship among the number of hidden layer neurons, training accuracy and time-consuming

| Experiment number | Number of hidden layer nodes | Accuracy | Training time | Encryption and classification run time |
|---|---|---|---|---|
| 1 | 10 | 90.1% | 143 s | 46.4 s |
| 2 | 20 | 90.19% | 247 s | 60.5 s |
| 3 | 40 | 90.24% | 400 s | 90.6 s |
| 4 | 80 | 90.23% | 698 s | 147.8 s |

From table 4, it can be seen that with the increasing of the number of neurons in the hidden layer, the training accuracy of neural network will increase at the beginning and then tend to flattened. Specially, when the number of neurons is 80, the training accuracy is slightly lower than that when the number of neurons is 40. However, as the number of neurons increases, the training time of the neural network and the classification time cost of the encrypted neural network will also rise. According to the above experimental analysis, the appropriate number of hidden layer neurons can be set as 20 in order to reduce the time cost without losing much precision.

There is also an empirical formula for the calculation of the number of hidden layer neurons.

$$m = \sqrt{n_i + n_o} + a \tag{8}$$

where $m$ is the number of hidden layer neurons, $n_i$ is the number of nodes in the input layer, $n_o$ is the number of nodes in the output layer, and $a$ is a constant between 1 and 10.

According to the empirical formula, the number of hidden layer neurons can be calculated and located at the interval [10, 20]. This is an approach to verify the rationality of the number of hidden layer neurons in our experiment.

The following problem is how to determine the learning rate of the neural network. The learning rate of neural networks $\eta$ is an important factor affecting the learning speed and accuracy of neural network. If the learning rate is too small, it will cause the neural network to converge slowly, which takes a lot of time. If the learning rate is too large, the magnitude of the parameter update will be very large, which will lead to the local optimum. Although the traditional stochastic gradient descent method (SGD) can help the neural network escape the local optimum, it never locates the minimum. Finally, we use an improved simulated annealing algorithm to solve this problem. The purpose of the algorithm is to gradually reduce the learning rate. The algorithm starts with a larger step size because it facilitates the rapid progress and avoids the local minimum, and then the step size becomes smaller and smaller to make the loss function be as close as possible to the global minimum. The specific algorithm is as follows:

**Algorithm 1 Update learning rate with improved simulated annealing algorithm**

```
Input：train_size: size of training set, batch_size: size of batch, iters_num: the number of iterations,
       eta0: initial learning rate.
Output：eta updated learning rate
eta0 ← 0.2
iter_per_epoch ← train_size / batch_size
epoch ← iters_num / iter_per_eopch
t0 ← eta0 * iter_per_epoch
t1 ← iter_per_epoch
for i = 0 to epoch do
    eta ← t0 / (i * 20 + t1)
    if  t < 0.011 then
        eta ← 0.01
    end if
end for
```

The number of training rounds (*epoch*) is determined by the size of the training set, the size of the batch and the number of iterations. When each *epoch* is completed, the learning rate is updated once. We assume that the number of iterations is 20000 and the initial learning rate is 0.2. When the number of trainings reaches 100 rounds, the learning rate will drop to around 0.01. In order to prevent the learning rate from being too low, it will be kept around 0.01. The above algorithm selects a relatively large learning rate in the early stage of training to enable the neural network to update parameters quickly, while it selects a relatively low learning rate for improving training accuracy in the later stage.

A simulated annealing algorithm is employed to gradually reduce the learning rate based on the initial value. Therefore, we set the learning rates to 0.5, 0.2, 0.1, and 0.05 respectively. The results are recorded in table 5.

**Table 5.** The relationship among neural network learning rate, training accuracy and time consumption

| Experiment number | Initial learning rate | Training time | Accuracy |
|---|---|---|---|
| 1 | 0.5 | 14.88s | 90.05% |
| 2 | 0.2 | 41.01s | 90.19% |
| 3 | 0.1 | 54.88s | 90.07% |
| 4 | 0.05 | 260.4s | 85.78% |

From table 5, it can be seen that when the initial learning rate is 0.5, the training time is the least, however the accuracy is not very high. When the learning rate is 0.05, the training time is the most, but the learning rate is low, resulting in a slow convergence rate. So in the case of 20,000 times iterations, the accuracy can only reach 85%. Therefore, we set the initial learning rate as 0.2.

A total of 12800 EEG samples are selected in the experiment, 80% of the samples are randomly selected as the training set, and the remaining 20% are used as the test set. The training set is used for parameter learning (weight and bias), and the test set is to verify the generalization ability of the model. Based on the above experiments and analysis, the best combination of parameters is obtained as shown in table 6.

**Table 6.** Optimal parameter setting

| Description | Symbol | Value |
|---|---|---|
| number of input layer nodes | $n_i$ | 44 |
| number of hidden layer nodes | $n_d$ | 20 |
| number of output layer nodes | $n_o$ | 4 |
| number of iterations | $iters_num$ | 20000 |
| size of batch | $b_s$ | 100 |
| initial learning rate | $\eta$ | 0.2 |

In the experiment, the neural network is trained three times using the relevant algorithms and the parameters in the table 6. The accuracy of the training set has been up to 90.37%, and the average accuracy has been 90.19%. The average accuracy of the test set has been 90.17%. And the training takes 250s time cost.

### 4.4 Model evaluation

The accuracy of the model cannot comprehensively evaluate the algorithm quality. For example, a simple cancer prediction function always returns a value of 0. If the accuracy of this prediction function is 99.5%, the error rate is 0.5%. For example, there is 0.5% of the real cancer, however it has always been misjudged. Therefore, we introduce a confusion matrix to evaluate the performance of the classifier. The confusion matrix can be defined as a situation analysis table that summarizes the prediction results of the classification model. Through the confusion_matrix() function in scikit-learn, we can get the following results as shown in table 7.

**Table 7.** Confusion matrix

| Truth / Prediction | eye opened | eye closed | both hands | both feet |
|---|---|---|---|---|
| eye opened | **620** | 65 | 4 | 13 |
| eye closed | 10 | **578** | 64 | 3 |
| both hands | 7 | 51 | **661** | 11 |
| both feet | 17 | 21 | 31 | **644** |

From table 7, we can see the difference between the actual label and the predicted label. For example, the number of samples is 620 when both the actual label and the predicted label are *eye opened*. If the actual label is *eye opened* and the predicted label is *eye closed*, the number of samples is 65. This also indicates that the correlation between *eye opened* and *eye closed* is high, leading to more mis-predictions. Oblique diagonally denotes the number of samples whose real category is consistent with the predicted category, and the value is also the largest, indicating that the performance of the model is satisfactory. Figure 5 shows the visualization of the confusion matrix.

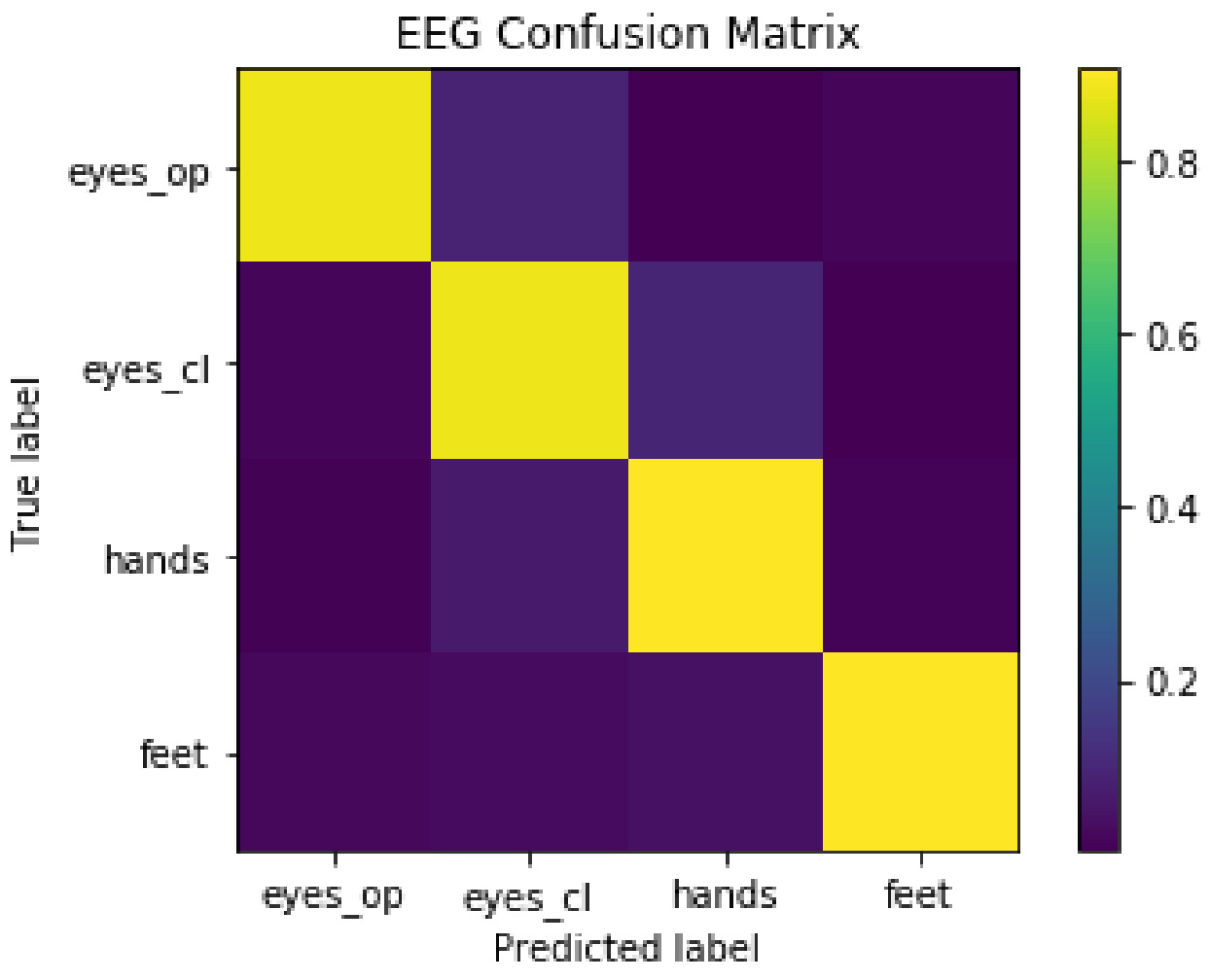

**Fig. 5.** Confusion matrix visualization

It can also be seen from the visualization of the confusion matrix that the diagonal color is yellowish, indicating that the probability that the actual label is consistent with the predicted label is above 90%, which further indicates that the model performance is desirable.

The effect can also be reflected intuitively through the precision rate and the recall rate. *Precision* refers to the proportion of all positive predictions that are correctly predicted. *Recall* refers to the proportion of all real positive observations that are correctly predicted. The specific formula is as follows

$$precision = \frac{TP}{TP + FP} \tag{9}$$

$$recall = \frac{TP}{TP + FN} \tag{10}$$

where *TP* is true positive, indicating that the ground truth is positive and the prediction is positive; *FP* is a false positive, indicating that the ground truth is negative but the prediction is positive; *FN* is false negative, indicating that the ground truth is positive but the prediction is negative.

For precision and recall, it can be combined into a simple metric called $F_1$-Score. $F_1$-Score is the weighted average of the precision and recall. The higher $F_1$-Score, the better the classification performance.

$$F_1 = \frac{2}{\frac{1}{precision} + \frac{1}{recall}} = 2\frac{precision * recall}{precision + recall} \tag{11}$$

Precision, recall and $F_1$-Score obtained from the confusion matrix are shown in table 8.

**Table 8.** Precision, Recall and $F_1$-Score

| Labels / Indicators | eye opened | eye closed | both hands | both feet | mean |
|---|---|---|---|---|---|
| precision | 94.80% | 80.83% | 86.98% | 95.98% | 89.65% |
| recall | 88.32% | 88.24% | 90.55% | 90.32% | 89.36% |
| $F_1$-Score | 91.45% | 84.38% | 88.72% | 93.06% | 89.40% |

From table 8, it can be intuitively seen that the effect of closing the eyes is the worst because the eye closed is more relevant to the eye opened. The reason for the good effect of the feet is that the feet are generally more sensitive. When the subject tightens the feet, the stimulation to the brain is higher than the other three actions, so the behavior of tightening the feet are easy to be distinguished. Moreover, the accuracy, recall and $F_1$-Scores in the experiments are close to 90%, which further demonstrates that our model has a satisfactory performance.

### 4.5 Time Cost Analysis

In order to test the time cost of our proposal, the experiments are performed on a PC using Intel i5-4210U processor, running 2.4GHz, 16GB RAM, and running Windows 10 operating system. Each piece of EEG data is composed of 44 features and one label. When we separately process one piece of EEG data, it takes about 45.7s, but as the amount of data processed increases, the average processing time of each piece of data will gradually decrease. We take 100 pieces of EEG data as an example, it can achieve hundreds of encryption and decryption operations per hour, and each piece of EEG data needs to run for about 20 seconds. Among them, it takes 12s to encrypt and encode each piece of data, 5s to network training, and 3s to decrypt and decode. These results are presented in table 9. It is worth to note that the EEG data set used in our experiment is non-linear and the structure is very complex. Therefore, it takes more processing time than ordinary data, but it still has an acceptant results.

**Table 9.** Average completion time (100 pieces of data)

| State | Time to compute |
|---|---|
| Encoding+Encryption | 12s |
| Network training | 5s |
| Decryption+Decoding | 3s |

### 4.6 Security Analysis

The homomorphic encryption operation of our proposal is based on the Paillier algorithm. Next, we mainly conduct security analysis from two aspects of algorithm security and privacy protection. First, from the perspective of algorithm security, the public key described in section 3.2 is $(n, g)$, where $g$ is an integer randomly selected from $Z_{n^2}^*$ and $n = pq$. The private key is $(\lambda, \mu)$, where $\lambda = lcm(p\text{-}1, q\text{-}1)$, and $\mu$ is the inverse of the modulo $n$ of $L(g^{\lambda} \bmod n^2)$. The sample data provided by the user is the ciphertext c. If the eavesdropper obtains the ciphertext c, where the public key is known and the private key is unknown, it cannot obtain the plaintext m through the tentative decryption operation. The safe of Paillier algorithm is based on the difficulty principle of large number decomposition that when the key setting is reasonable, $p$ and $q$ derived by the public key $n$ is not feasible in time. Under the premise, the attacker cannot calculate $g^{\lambda}$ and get the private key $\mu$. Therefore, even if the ciphertext is obtained, the plaintext cannot be cracked, which proves the security of the algorithm.

On the other hand, from the perspective of privacy in the application scenario, our experiments mainly include offline and online parts (see Figure 1). The offline part is mainly used for model training to determine hyperparameters. The data sets used to training may not involve user privacy. The online part is mainly used for model prediction, and the data here is operated with the format of ciphertext. Our solution can allow data owners to send their EEG data to the prediction service provider with the format of ciphertext. The prediction service provider can use the encrypted EEG data to complete the prediction and return the result to a key owner who can decrypt them. This process uses a public key for encryption and a private key for decryption. Among them, encryption guarantees the confidentiality of data, and the prediction result is also in the ciphertext form. The prediction service provider cannot obtain the private key, so it can neither decrypt the received data nor the prediction result. This means the prediction service provider cannot obtain any information about users. Therefore, it also proves the privacy of our method.

### 4.7 Performance Comparison

The use of the public datasets just provided by PhysioNet for experimental analysis is not comprehensive. In order to further demonstrate the feasibility, our proposal will be applied to other EEG-related datasets, where the names of experimental datasets and specific parameters are shown in table 10 together with PhysioNet datasets.

**Table 10.** Datasets and parameters

| Datasets | Sample | channels | Class | Epochs | Learning rate |
|---|---|---|---|---|---|
| PhysioNet | 12800 | 44 | 4 | 100 | 0.2 |
| BCI Competition IV | 10000 | 25 | 4 | 100 | 0.2 |
| EPILEPSIAE | 8000 | 122 | 2 | 80 | 0.2 |

In table 10, the second dataset is mainly used for the classification of motor imaging tasks, which can identify left hand, right hand, foot and tongue, respectively. The EPILEPSIE database is by far the largest and most comprehensive database for human surface and intracranial EEG data, which is mainly used to predict whether or not suffering from epilepsy.

We use the following formula to calculate the error rate as a cost of privacy protection.

$$Error\ rate\ =\ \frac{number\ of\ test\ samples\ misclassified}{number\ of\ training\ samples} \tag{12}$$

The experimental results obtained according to formula 12 and the parameters in table 10 are shown in table 11.

**Table 11.** Error rates comparison

| Datasets | Non-privacy preserving (%) | Privacy preserving (%) |
|---|---|---|
| PhysioNet | 5.00% | 10.00% |
| BCI Competition IV | 4.33% | 9.67% |
| EPILEPSIAE | 9.50 | 14.83% |

In table 12, the metric of non-privacy preserving indicates the EEG data is unencrypted while that of privacy preserving represents the EEG data is encrypted. It can also be seen from table 12 that the loss of accuracy (i.e. the difference between the two kinds of error rates) is around 5%. The EPILEPSIAE dataset may have a higher error rate due to the use of more channels. This demonstrates that our proposal has a satisfactory feasibility owed to the minor difference of classification accuracy between encrypted data and unencrypted data.

Paillier+FNN method is adopted as the classifier to classify encrypted EEG data. To demonstrate the efficiency of this method, we compare it with several widely used classification methods. All classifiers work on the same EEG dataset, and the experimental results are shown in table 12 and table 13, respectively.

**Table 12.** Performance comparison (PhysioNet)

| Author | Classifier | Feature extraction | Encryption method | Binary/Multi | Accuracy |
|---|---|---|---|---|---|
| Alomari[11] | SVM+NN | DWT | — | Binary | 74.97% |
| Shenoy[12] | SVM | FBCSP | — | Binary | 82.06% |
| Ward[13] | UBM | JFA | — | Multi(3) | 80% |
| Gilad[17] | CNN | — | FHE | Multi(3) | 90% |
| Wu [18] | KNN | — | Paillier | Multi(4) | 87.35% |
| **Ours** | **FNN** | — | **Paillier** | **Multi(4)** | **90.17%** |

**Table 13.** Performance comparison (BCI Competition IV)

| Author | Classifier | Feature extraction | Encryption method | Binary/Multi | Accuracy |
|---|---|---|---|---|---|
| Alomari[11] | SVM+NN | DWT | — | Binary | 76.23% |
| Shenoy[12] | SVM | FBCSP | — | Binary | 83.19% |
| Ward[13] | UBM | JFA | — | Multi(3) | 81.93% |
| Gilad[17] | CNN | — | FHE | Multi(3) | 90.57% |
| Wu [18] | KNN | — | Paillier | Multi(4) | 88.16% |
| **Ours** | **FNN** | — | **Paillier** | **Multi(4)** | **91.24%** |

In table 12 and table 13, our experiments are conducted on the datasets provided by PhysioNet and BCI Competition, respectively. SVM denotes Support Vector Machine; NN denotes Neural Network; DWT denotes Discrete Wavelet Transform; FBCSP denotes Filter Bank Co-Space Mode; UBM denotes Universal Background Mode; JFA denotes Joint Factor Analysis; CNN denotes Convolutional Neural Network; FHE denotes Fully Homomorphic Encryption Algorithm; KNN denotes K-Nearest Neighbor.

From table 12 and table 13, it can be seen that the traditional machine learning all require artificial feature extraction, and they cannot achieve the satisfactory accuracy, which also proves the rationality of our choice of feed-forward neural network. The method proposed by Gilad et al. [17] is applied to the experimental dataset, whose accuracy is slightly lower than ours, however this method uses the fully homomorphic encryption algorithm. Under the same conditions, the full homomorphic encryption algorithm will have more time consumptions compared to the semi-homomorphic encryption algorithm. This is exactly why we choose the Paillier algorithm. Although the scheme of Wu et al. [18] also adopted the Paillier algorithm, the use of KNN classifier causes that its accuracy is lower than that of our scheme. Since the number of channels and samples of the datasets in table 13 is smaller than those in table 12, the accuracy of table 13 is lower. In conclusion, our proposal has higher accuracy and lower time complexity than other methods.

## 5 Conclusion

In this paper, we proposed an improved neural network model for processing multi-class recognition of encrypted EEG signals. In order to prevent the user's sensitive information from leaking and curb illegal operations, the Paillier algorithm was used to encrypt the EEG data, which was converted to integers through appropriate scaling to implement floating-point operations. Compared with the fully homomorphic encryption algorithm, the time complexity was further reduced. In order to maintain a certain homomorphism during the data processing, an improved feed-forward neural network model was adopted for training, and an approximate function was used instead of the activation function. Compared with traditional machine learning techniques, neural network algorithms did not require artificial feature extraction, and further improved the accuracy and stability of EEG classification. In addition, the confusion matrix is employed to verify and evaluate the proposed model and method on the public EEG datasets, and the accuracy of multi-class EEG recognition is improved to more than 90%. The theoretical analysis and experimental results show that our proposal has the satisfactory accuracy, efficiency and security.

Although our proposal has more satisfactory results compared with other solutions, it still has some unsolved challenging problems, such as the more stringent accuracy requirements in telemedicine scenario, the lower time complexity especially for the large volume datasets, etc. Therefore, future work needs to be studied from the following three aspects: 1) Seek alternative solution of activation function to further improve the accuracy of classification; 2) Based on a large scale EEG data set, analyze and study deeper neural network models to achieve a better trade-off between accuracy and efficiency; 3) Find more efficient coding schemes to allow smaller parameters to achieve faster homomorphic calculations.

## Reference

[1] Birbaumer N, Cohen L G. Brain-computer interfaces: communication and restoration of movement in paralysis [J]. Journal of Physiology (Oxford), 2007, 579(3):621-636.

[2] Varkuti B, Guan C, Pan Y, et al. Resting State Changes in Functional Connectivity Correlate With Movement Recovery for BCI and Robot-Assisted Upper-Extremity Training After Stroke [J]. Neurorehabilitation and Neural Repair, 2013, 27(1):53-62.

[3] McCane L M, Heckman S M, McFarland D J, et al. P300-based brain-computer interface (BCI) event-related potentials (ERPs): People with amyotrophic lateral sclerosis (ALS) vs. age-matched controls [J]. Clinical Neurophysiology, 2015, 126(11): 2124-2131.

[4] Zhang X, Yao L, Huang C, et al. Enhancing Mind Controlled Smart Living Through Recurrent Neural

Networks [C]//Proceedings of the 23rd ACM SIGKDD International Conference on Knowledge Discovery and Data Mining, Halifax, NS, Canada, 13-17 Aug. 2017: 43-52.

[5] Stefano Filho C A, Attux R, Castellano G. Can graph metrics be used for EEG-BCIs based on hand motor imagery [J]. Biomedical Signal Processing and Control, 2018, 40: 359-365.

[6] Jiao Y, Zhang Y, Chen X, et al. Sparse group representation model for motor imagery EEG classification [J]. IEEE journal of biomedical and health informatics, 2018, 23(2): 631-641.

[7] Chatterjee R, Moitra T, Islam S K H, et al. A novel machine learning based feature selection for motor imagery EEG signal classification in internet of medical things environment [J]. Future Generation Computer Systems, 2019, 98: 419-434.

[8] Tang Z, Li C, Sun S. Single-trial EEG classification of motor imagery using deep convolutional neural networks [J]. Optik-International Journal for Light and Electron Optics, 2017, 130: 11-18.

[9] Li X, La R, Wang Y, et al. EEG-based mild depression recognition using convolutional neural network [J]. Medical and Biological Engineering and Computing, 2019, 57: 1341-1352.

[10] Nienhold D, Schwab K, Dornberger R, et al. Effects of Weight Initialization in a Feedforward Neural Network for Classification Using a Modified Genetic Algorithm [C]//2015 3rd International Symposium on Computational and Business Intelligence (ISCBI), Bali, Indonesia, 7-9 Dec. 2015: 6-12.

[11] Alomari M H, AbuBaker A, Turani A, et al. EEG mouse: A machine learning-based brain computer interface [J]. International Journal of Advanced Computer Science and Applications, 2014, 5(4): 193-198.

[12] Shenoy H V, Vinod A P, Guan C. Shrinkage estimator based regularization for EEG motor imagery classification [C]//2015 10th International Conference on Information, Communications and Signal Processing (ICICS), Singapore, 2-4 Dec. 2015: 1-5.

[13] Ward C, Picone J, Obeid I. Applications of UBMs and I-vectors in EEG subject verification [C]//2016 38th Annual International Conference of the IEEE Engineering in Medicine and Biology Society (EMBC), Orlando, FL, USA, 17-20 Aug. 2016: 748-751.

[14] Kumar P, Singh M D, Saxena A. HEMIN: A Cryptographic Approach for Private k-NN Classification [C]// Proceedings of The 2008 International Conference on Data Mining, Las Vegas, USA, 5-7 Jan. 2008: 500-505.

[15] Clifton C, Kantarcioglu M, Vaidya J, et al. Tools for privacy preserving distributed data mining [J]. ACM Sigkdd Explorations Newsletter, 2002, 4(2): 28-34.

[16] Chabanne H, de Wargny A, Milgram J, et al. Privacy-preserving classification on deep neural network [J]. IACR Cryptology ePrint Archive, 2017, 2017: 35.

[17] Gilad-Bachrach R, Dowlin N, Laine K, et al. Cryptonets: Applying neural networks to encrypted data with high throughput and accuracy [C]// Proceedings of the 33rd International Conference on Machine Learning, New York, NY, USA, 19-24 Jun. 2016: 201-210.

[18] Wu W , Parampalli U , Liu J , et al. Privacy preserving k-nearest neighbor classification over encrypted database in outsourced cloud environments [J]. World Wide Web, 2018(11):1-23.

[19] Marcano, Nestor Javier Hernandez, et al. On Fully Homomorphic Encryption for Privacy-Preserving Deep Learning [C] // 2019 IEEE Global Communications Conference, Waikoloa, HI, USA, 9-13 Dec. 2019: 1-6.

[20] Weiru W, Yanfen G, Chi-Man V, et al. Homo-ELM: fully homomorphic extreme learning machine [J]. International Journal of Machine Learning and Cybernetics, 2020, 1: 1-10.

[21] Paillier P. Public-key cryptosystems based on composite degree residuosity classes [C]//International Conference on the Theory and Applications of Cryptographic Techniques. Springer, Berlin, Heidelberg, 1999: 223-238.

[22] Xie P, Bilenko M, Finley T, et al. Crypto-nets: Neural networks over encrypted data [R]. arXiv preprint arXiv:1412.6181, 2014.

[23] Schalk G, McFarland D J, Hinterberger T, et al. BCI2000: a general-purpose brain-computer interface (BCI) system [J]. IEEE Transactions on biomedical engineering, 2004, 51(6): 1034-1043.